\documentclass[runningheads]{llncs}

\usepackage{eccv}

\usepackage{eccvabbrv}

\usepackage{graphicx}
\usepackage{booktabs}
\usepackage{colortbl}
\usepackage{multirow}
\usepackage{floatrow}
\usepackage{sidecap}
\usepackage[export]{adjustbox}
\usepackage{dcolumn}

\usepackage{pifont}%
\newcommand{\cmark}{\ding{51}}%
\newcommand{\xmark}{\ding{55}}%

\newcolumntype{d}[1]{D{.}{.}{#1}}
\def\modelName{NMS\xspace}
\definecolor{aoenglish}{rgb}{0.0, 0.7, 0.0} %

\definecolor{htmlcssgreen}{rgb}{0.0, 0.5, 0.0}

\newcommand{\increase}[1]{\scriptsize \textcolor{aoenglish}{$\uparrow$#1} \normalsize}
\newcommand{\decrease}[1]{\scriptsize \textcolor{red}{$\downarrow$#1} \normalsize}
\newcommand{\supp}[1]{{\color{blue}  #1}}

\usepackage[accsupp]{axessibility}  %

\usepackage{hyperref}

\usepackage{orcidlink}

\begin{document}

\title{Sync from the Sea: Retrieving Alignable Videos from Large-Scale Datasets}

\titlerunning{Sync from the Sea: Retrieving Alignable Videos from Large-Scale Datasets}

\author{Ishan Rajendrakumar Dave\inst{1}\thanks{Majority of work done as an intern at Adobe Research, USA}\orcidlink{0000-0001-9920-6970} \and  Fabian Caba Heilbron\inst{2}\orcidlink{0000-0002-3129-1985} \and Mubarak Shah\inst{1}\orcidlink{0000-0001-6172-5572} \and Simon Jenni\inst{2}\orcidlink{0000-0002-9472-0425}}

\authorrunning{Dave et al.}

\institute{Center for Research in Computer Vision,
University of Central Florida, USA \and
Adobe Research, USA\\
\email{ishandave@ucf.edu}, \email{caba@adobe.com}, \email{shah@crcv.ucf.edu}, \email{jenni@adobe.com}\\
\url{https://daveishan.github.io/avr-webpage/}
}

\maketitle

\begin{abstract}
Temporal video alignment aims to synchronize the key events like object interactions or action phase transitions in two videos. 
Such methods could benefit various video editing, processing, and understanding tasks. 
However, existing approaches operate under the restrictive assumption that a suitable video pair for alignment is given, significantly limiting their broader applicability. 
To address this, we re-pose temporal alignment as a search problem and introduce the task of Alignable Video Retrieval (AVR). 
Given a query video, our approach can identify well-alignable videos from a large collection of clips and temporally synchronize them to the query. 
To achieve this, we make three key contributions: 
1) we introduce DRAQ, a video alignability indicator to identify and re-rank the best alignable video from a set of candidates; 
2) we propose an effective and generalizable frame-level video feature design to improve the alignment performance of several off-the-shelf feature representations, and 
3) we propose a novel benchmark and evaluation protocol for AVR using cycle-consistency metrics.
Our experiments on 3 datasets, including large-scale Kinetics700, demonstrate the effectiveness of our approach in identifying alignable video pairs from diverse datasets.

\end{abstract}

\section{Introduction}

\begin{figure}[t]
    \centering
    \begin{subfigure}[b]{0.39\textwidth}
        \includegraphics[width=\textwidth]{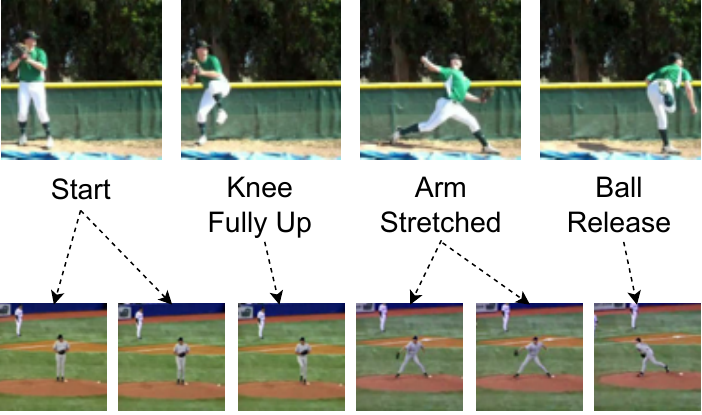}
        \caption{\textbf{Regular Temporal Alignment}: A pair of videos from the same action class is given. The goal is to align them, \ie, match their key-event frames}
        \label{fig:sub1}
    \end{subfigure}
    \hfill
    \begin{subfigure}[b]{0.58\textwidth}
        \includegraphics[width=\textwidth]{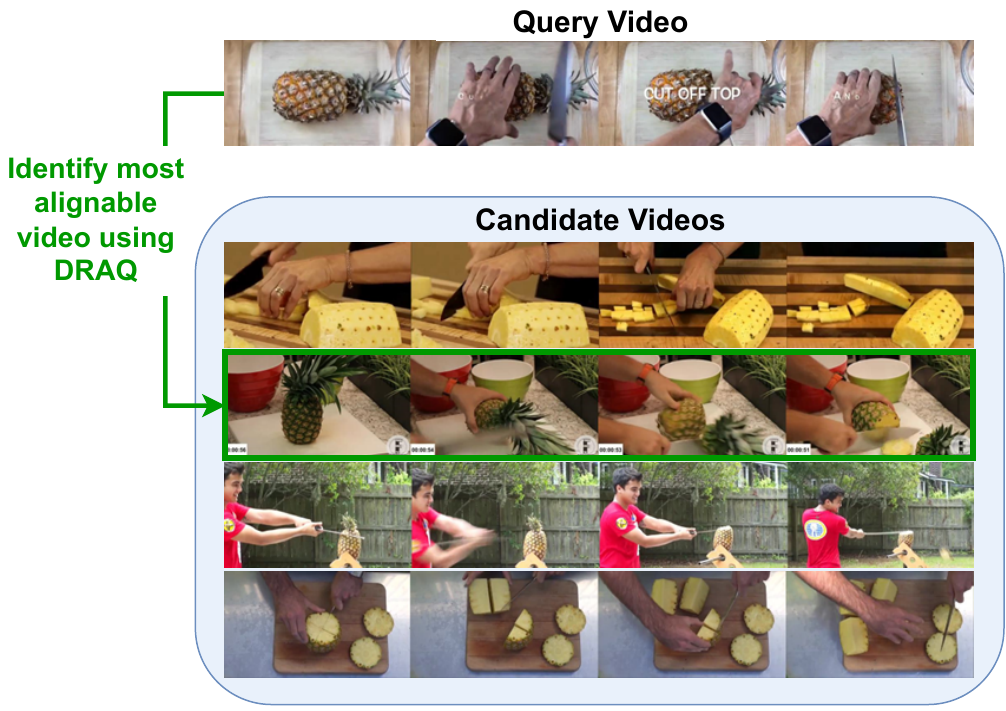}
        \caption{\textbf{Proposed Alignable Video Retrieval (AVR)}:
Given a query video, the goal is to find the best alignable video from candidate videos of the video search results.}
        \label{fig:sub2}
    \end{subfigure}
    \caption{ \textbf{Alignable Video Retrieval.}
    While some actions, like "baseball swing" (left), permit temporal alignment in virtually all cases due to their fixed sequence of action phases, general videos from other action classes, like "cutting pineapple" (right), exhibit much more variability. 
    Knowledge of the action category alone is insufficient to identify alignable pairs for these cases, and a deeper temporal understanding of the videos is required to identify alignable videos. 
    We propose DRAQ, an alignability score that can reliably identify the alignable video pair (red) among the set of candidates. }
    \label{fig:motivation}
\end{figure}

Video understanding has made great progress in recent years, as evidenced by numerous tasks and benchmarks ranging from action recognition~\cite{ucf101,kay2017kinetics} and localization~\cite{caba2015activitynet,gu2018ava} to video editing \cite{ceylan2023pix2video} and generation \cite{singer2022make}.
A key challenge in the semantic and temporal understanding of videos is that of temporal video alignment. 
We say that two videos are temporally aligned when their key events (\eg, action phase transitions, object interactions, etc.) co-occur exactly. 
For example, given two videos showing a "baseball swing," a video alignment approach would thus warp the videos so that the moment the person starts the swing and the moment the ball is released happen simultaneously in the two videos (see Figure~\ref{fig:sub1}). 
Such methods can enable various applications in video processing and analysis. 
For example, it enables example-based video retiming \cite{jenni2022video} where a given video is warped according to dynamics found in another video, automated video clip replacement without the need for editing (\eg, when license issues prevent the use of the original clip), the automatic time-aligned transfer of audio tracks \cite{dwibedi2019temporal} or video effects between clips, to name a few.

However, existing approaches for video alignment \cite{dwibedi2019temporal,haresh2021learning,chen2022frame} primarily consider a restrictive, weakly supervised setting, where the pair of videos for alignment is assumed given, and only the alignment of the two videos is to be found. 
Because identifying alignable videos is a challenging problem that has remained unaddressed thus far, this limits the broader applicability of video alignment methods.
Furthermore, existing methods train and test predominantly on a limited number of well-behaved action categories with well-delignated action phases.
General real-world videos are often not as well-behaved as the types of videos considered in existing video alignment benchmarks. 
For example, they might not conform to a fixed sequence of key events and might exhibit large variations in how an action is performed.
As illustrated in Figure~\ref{fig:motivation}, while videos of actions like "baseball swing" are almost always alignable due to their shared sequence of action phases, for videos of more general categories like "cutting pineapple," the larger variation in videos makes knowledge of the action category insufficient for identifying alignable video pairs. 
Given these observations, off-the-shelf video retrieval methods (\eg, trained for action recognition) by themselves are insufficient for identifying alignable video pairs and a tailored solution for filtering or re-ranking candidate pairs is required. 
Therefore, in this paper, we address the question: \emph{How to identify alignable videos from a large collection?}

Toward this goal, we introduce the task of Alignable Video Retrieval (AVR), tackling the thus far unaddressed problem of \emph{identifying} alignable videos from a large dataset. 
The AVR task naturally combines the fundamental video understanding problems of retrieval and alignment, and we make several technical contributions toward a solution. 
As a first contribution, we propose a method for identifying the best alignable clips from a set of retrieved candidate videos. 
To this end, we introduce an alignability indicator that scores how alignable two video clips are. 
Our method, Dynamic Relative Alignment Quality (DRAQ), compares the optimal alignment cost (as obtained through Dynamic Time Warping \cite{muller2007dynamic}) to the average cost of multiple random sub-optimal alignments. 
Experimentally, we show that this intuitive indicator effectively identifies the best alignable videos by showing its high correlation with the action phase agreement of aligned videos and performance improvements in reranking video search results. 
As a second contribution, we propose a method to effectively improve the performance of off-the-shelf video-frame representations for video alignment. 
To this end, we introduce a feature contextualization approach, which augments a given frame-level feature representation with additional temporal context.
Experimentally, we show that various off-the-shelf representations benefit from such contextualization for video alignment. 
Finally, as our third contribution, we propose a set of benchmarks and evaluation protocols to measure AVR performance. 
Our evaluation includes existing datasets with dense action phase labels where we propose aligned phase agreement to measure alignment quality and newly annotated Kinetics videos to evaluate the full AVR pipeline in a cycle-consist manner on a more diverse set of natural videos.

\section{Prior Work}

Our work on Alignable Video Retrieval (AVR) relates to several topics in computer vision, including video alignment, retrieval, and temporal feature learning.

\noindent \textbf{Video Alignment.}
Several recent works proposed targeted training strategies to learn video representations for alignment. 
Many of these works study a weakly supervised setting, where video pairs showing the same action are used for learning \cite{dwibedi2019temporal,haresh2021learning,fakhfour2023video,zhang2023modeling}.
These methods rely on cycle consistency \cite{dwibedi2019temporal}, DTW-based temporal alignment constraints \cite{haresh2021learning,fakhfour2023video}, or a combination of the two \cite{hadji2021representation}.
Another line of work focuses on purely unsupervised learning of frame-level features for alignment on unlabelled videos. 
These works build on variations of frame-contrastive learning objectives on augmented video clips \cite{chen2022frame, dave2024finepseudo}, some also incorporating weak supervision \cite{zhang2023modeling}.
Alignment with the resulting frame-level features is typically performed via Dynamic Time Warping (DTW)~\cite{muller2007dynamic}. Some methods also leverage differentiable formulations thereof \cite{cuturi2017soft}, and variants robust to outliers have been proposed \cite{dropdtw}.
Instead of designing a novel learning strategy for video alignment, we build our approach on existing pre-trained video representations, for which we introduce a video contextualization procedure to enhance their alignment capabilities.

\noindent \textbf{Video Retrieval.} 
A comprehensive AVR solution relies on good candidate proposals obtained through large-scale video-to-video retrieval.
While early video search approaches were based on hand-crafted features \cite{sivic2006video}, recent methods use neural-network representations, \eg, learned through action recognition \cite{feichtenhofer2019slowfast,tran2018closer, gabv2} or self-supervised learning \cite{qian2021spatiotemporal, tclr, jenni2021time, thoker2023tubelet, jenni2023audio}.
Of particular interest are visual retrieval methods that also aim at localizing a matching segment in a video. 
Such approaches are prevalent in the video copy detection literature \cite{baraldi2018lamv,jiang2016partial,Douze2015CirculantTE,tan2022fast,dropdtw,han2021video,transvcl}.
Most related are two-stage approaches \cite{tan2022fast,black2023vader}, where an initial coarse retrieval and localization are then refined with an "alignment" stage. 
Video copy detection approaches have in common that they are looking for identical video segments (up to video processing artifacts). Instead, our goal is to retrieve videos that permit a semantic, temporal alignment between different videos.

\noindent \textbf{Temporal Video Representation Learning.} 
A key requirement for accurate video alignment is a per-frame representation that captures the temporal features in a video and can discriminate subtle changes as a scene evolves over time.
Several works explored the use of temporal self-supervision to learn such temporal sensitivity, \eg, through pretext tasks about the ordering of video frames \cite{misra2016shuffle,fernando2017self,lee2017unsupervised}, or the classification of playback direction \cite{wei2018learning} and speed \cite{epstein2020oops,benaim2020speednet, simon}. 
We leverage video features learned through such temporal self-supervision \cite{vtn_ssl} both for retrieval and alignment in our approach.

\section{Temporally Aligned Video Retrieval}
\label{sec:method}

\begin{figure*}[t]
    \centering
    \includegraphics[width=\linewidth]{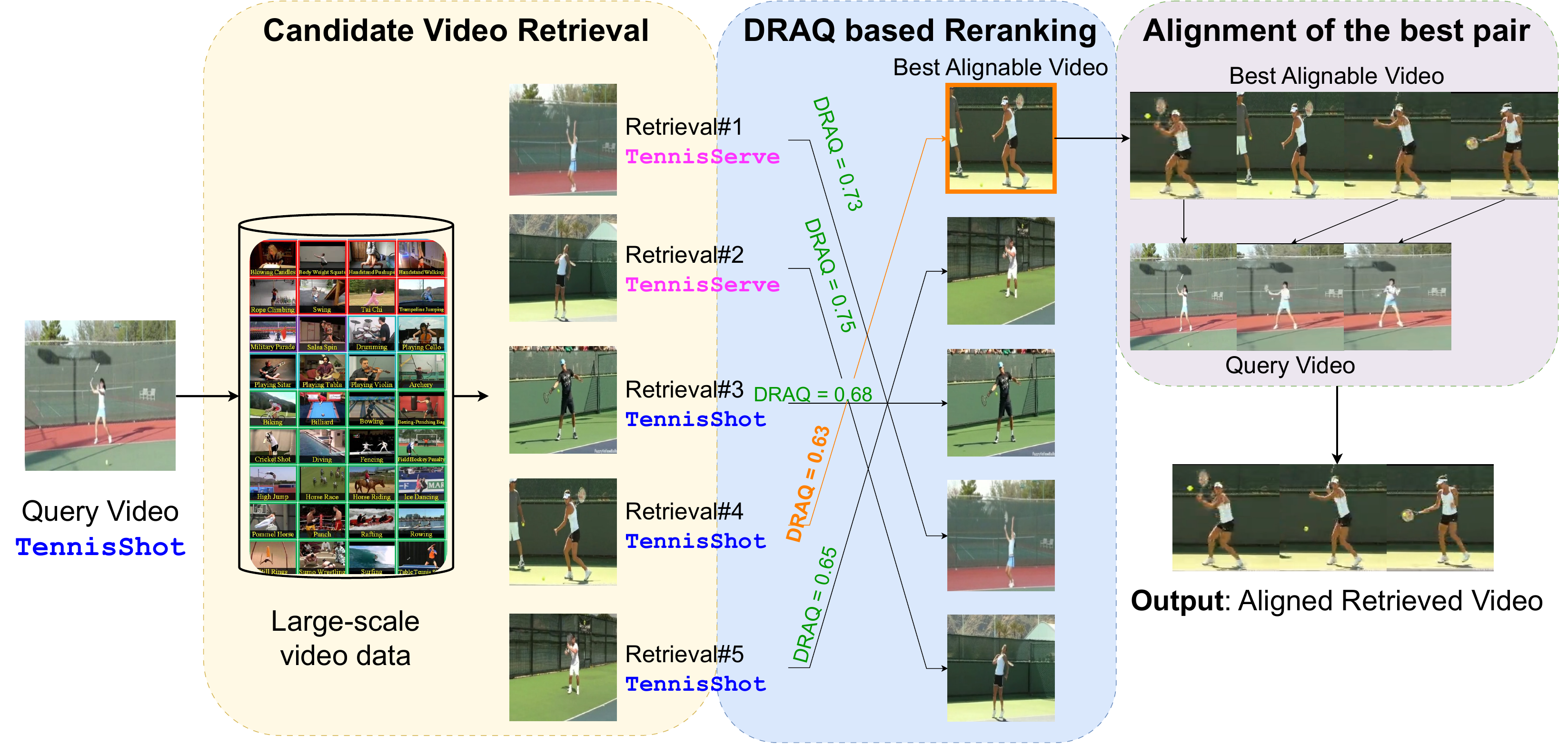}
    \caption{\textbf{Model Overview.}
    We introduce a model for Aligned Video Retrieval (AVR): Given an input query video clip, our model aims to find and temporally align the best matching video among a large collection of videos. 
    Our approach has three stages: 1) candidate retrieval from a large-scale database, 2) re-ranking of the top candidates to identify the most alignable clip using our procedure DRAQ, and 3) alignment of query and top match using DTW on our contextualized frame-level features. 
    }
    \label{fig:overview}
\end{figure*}

We propose a solution for temporally Aligned Video Retrieval (AVR). 
Given a query video clip, our approach aims to identify the best alignable clip among a large collection. It then temporally aligns this best-alignable clip to the query, accurately transferring the timing of key events in the query clip.
Our approach for retrieval and alignment of videos consists of three stages: 1) large-scale retrieval of candidate video clips from a large collection, 2) re-ranking of candidates based on a novel alignability indicator DRAQ, and 3) aligning of the best pair using DTW on contextualized frame-level features.
An overview of our system is provided in Figure~\ref{fig:overview}.

We build on video representations obtained with temporal self-supervised pre-training as described in \cite{vtn_ssl} to represent whole video clips and their frames.
These representations are shown in \cite{vtn_ssl} to achieve state-of-the-art results in video action retrieval, and as our experiments in Section~\ref{sec:exp} show, they also achieve state-of-the-art in video alignment provided an additional global video contextualizing of the per-frame features we introduce.
Given a video $V_i \in \mathbb{R}^{T \times H \times W \times 3}$ consisting of $T$ frames of size $H\times W$, we encode it with the encoder $E$ from \cite{vtn_ssl} to obtain the feature sequence
\begin{equation}
    \label{eq:feats}
    F_i=[f_1^{(i)}, \ldots, f_T^{(i)}] \in \mathbb{R}^{T \times d},
\end{equation}
where $d$ is the size of each per-frame feature vector $f_j^{(i)}$.

\noindent \textbf{Candidate Video Retrieval.}
To retrieve candidate video clips from a large collection of clips, we build a search index using temporally aggregated feature vectors $\Bar{F_i} = \frac{1}{T}\sum_{j=1}^T f_j^{(i)} $.
These clip-level feature vectors are stored in an efficient approximate nearest-neighbor data structure.
As an additional preprocessing step, we standardize the feature vectors based on the mean and standard deviation computed on the retrieval dataset. 
Given a query video, we use cosine similarity to find the top-$k$ candidates for alignment from the dataset.

\noindent \textbf{Contextualized Frame-Level Features for Alignment.}
Given two videos, $ V_1 $ and $ V_2 $, with $ n $ and $ m $ frames, respectively, we now describe how to construct contextualized features for each frame that will be used for temporal alignment.  
We augment the base frame-level features $f_i^{(j)}$ from $E$ (see Equation~\ref{eq:feats}) with sequence-level context to better support the alignment task.
Concretely, each feature should not just capture scene features at a specific moment (\eg, the pose of a person at some point in an action sequence) but also how that moment fits into the overall action sequence (\eg, is it at the end or beginning). 
To endow the video features with such temporal context, we concatenate them with the cumulative sum of features up to each time step. 
Concretely, our contextualized features for a video with $T$ frames are given by
\begin{equation}
    \Bar{f}_j^{(i)} = f_j^{(i)} \oplus \frac{1}{T} \sum_{t=1}^j f_t^{(i)} \in \mathbb{R}^{T \times 2d},
\end{equation}
where $\oplus$ indicates concatenation along the channel dimension.
Finally, we standardize the frame features per clip via zero-centering, \ie, working with
\begin{equation}\label{eq:frame_feats}
    \hat{f}_i^{(1)} =  \Bar{f}_i^{(1)} - \frac{1}{T} \sum_{l=1}^T \Bar{f}_l^{(1)},
\end{equation}
instead of $ f_i^{(1)}$ in the following ($\hat{f}_j^{(2)}$ is defined similarly).
Note that this approach is very general and can be applied to any frame-level representation and video length, even when the sequence length at inference time is very different from pre-training (as is the case with our default video features).

\noindent \textbf{Temporal Alignment via DTW.}
Given pairs of candidate video clips from our retrieval stage, we leverage Dynamic Time Warping (DTW) to find an optimal alignment between the two frame sequences.
Since our alignability indicator (DRAQ) is closely related to the computations required for DTW, we provide a detailed description of DTW first.
DTW operates on a cost matrix $C \in \mathbb{R}^{n \times m }$, which quantifies the similarity between feature vectors from the two videos.
To compute $ C $, we employ the frame-level distance
\begin{equation}
C(i,j) = 1 - \frac{\hat{f}_i^{(1)} \cdot \hat{f}_j^{(2)}}{\lVert \hat{f}_i^{(1)} \rVert \lVert \hat{f}_j^{(2)} \rVert},
\end{equation}
where $\hat{f}_i^{(1)} \cdot \hat{f}_j^{(2)}$ denotes the dot product between the two feature vectors, and $ \lVert \cdot \rVert $ is the vector norm.
DTW then determines the optimal alignment path $ P_{\text{DTW}} $, which minimizes the cumulative cost through $C$ from the top-left $(1, 1)$ to the bottom right $(n,m)$. 
More concretely, a path $P$ through a cost matrix $C$ of size $n \times m$ is defined as a sequence of tuples
\[ P = \left( (i_1, j_1), (i_2, j_2), \ldots, (i_L, j_L) \right), \]
where:
\begin{itemize}
    \item $ i_1 \leq i_2 \leq \ldots \leq i_L $ and  $ j_1  \leq j_2 < \ldots  \leq j_L$
    \item $L$ is the length of the path with $L \leq n + m$.
    \item The start and end points are fixed: $(i_1, j_1) = (1,1)$ and $(i_L, j_L) = (n,m)$.
\end{itemize}

To compute the optimal path $ P_{\text{DTW}} $, let $ D $ be a matrix of the same dimensions as $ C $ where each entry $ D(i,j) $ represents the minimum cumulative cost of aligning the sequences up to $ \hat{f}_i^{(1)} $ and $ \hat{f}_j^{(2)} $. This is computed recursively as
\begin{equation}
D(i,j) = C(i,j) + \min_{(u,v) \in \Delta} D(i-u, j-v),
\end{equation}
where $\Delta = \{ (0,1), (1,0), (1,1) \}$ is a set of offsets corresponding to the three valid moves.

The optimal path $ P_{\text{DTW}} $ is then traced back from $ D(n,m) $ to $ D(1,1) $, choosing at each step the direction that resulted in the minimal cumulative cost.
This path then also defines the optimal temporal alignment between the two frame sequences in our method. 
In some cases, we might want to restrict ourselves to solutions that keep one of the involved frame sequences unwarped, \eg, in example-based video re-timing or similar applications. 
We opt to skip any still frames for that particular sequence in the optimal path for such cases (\ie, we delete such index tuples from the path).

\noindent \textbf{DRAQ: Assessing Alignability for Re-Ranking.}
For our approach to AVR to work, it is crucial to have a way to assess the quality of an alignment between two videos. 
A straightforward choice would be to use the optimal DTW path cost $D(n,m)$. 
However, $D(n, m)$ builds on the absolute similarity of frames between the two sequences, which is more strongly influenced by the appearance of the two clips rather than their temporal alignability. 

Therefore, we propose a new method to assess the quality of an alignment between two videos, Dynamic Relative Alignment Quality (DRAQ).
The idea behind DRAQ is to compare the optimal alignment (as obtained with DTW) to an average random alignment between the two videos. 
Intuitively, if the optimal alignment achieves a clearly lower cost than a random alignment, we can be more confident that a meaningful synchronization of key video moments could be achieved.
To determine this baseline cost of a random alignment, we generate $ k $ random alignment paths in the cost matrix $ C $.
To generate a random path, we start at $ (i, j) = (n,m) $ and, at each step, select one of the possible moves $(\delta_i, \delta_j)\in \{-1, 0\}^2$ as follows:
\begin{itemize}
    \item Sample  $\delta_i=-1$ with probability $P_{\text{up}}=\frac{i}{i+j}$        \item Sample  $\delta_j=-1$ with probability $P_{\text{left}}=\frac{j}{i+j}$
    \item Move into direction $(\delta_i, \delta_j)$ until $(1, 1)$ is reached
\end{itemize}
We ignore any steps equalling $(\delta_i, \delta_j)=(0, 0)$ in this process.
Note how the paths are being biased towards the top-left direction by making the direction probabilities proportional to $i$ and $j$. 
This is important to make the random paths more "challenging" compared to completely random sampling. 
To compute the cost of a randomly sampled path, we sum up all the corresponding entries in $C$.
This process is repeated $ k $ times to generate $ k $ random path costs. 
The costs of the $k$ random paths are finally averaged to obtain $ \text{Cost}_{\text{random}} $.

The DRAQ metric is then defined as the ratio of the cumulative cost along the optimal alignment path to the average cost of $ k $ random paths, \ie,
\begin{equation}
\text{DRAQ} =  \frac{D(n,m)}{\text{Cost}_{\text{random}}},
\end{equation}
where $ D(n,m) $ is the cumulative cost of the optimal alignment path.
Because $C$ is computed only once and sampling random paths through $C$ is very efficient, DRAQ has minimal computational overhead compared to DTW.

With this formulation, the DRAQ score provides a robust mechanism to quantify video alignment quality by comparing the efficacy of a globally optimal alignment scheme (DTW) to average random alignments.
Furthermore, because DRAQ is defined as a ratio of path costs, it does not suffer from the same appearance bias as DTW and serves as a better alignability indicator, as demonstrated in our experiments.
\\

\noindent \textbf{The Aligned Video Retrieval Pipeline.}
To summarize, given a query video, we build an AVR pipeline consisting of 1) candidate alignable video retrieval using $k$-nearest neighbor retrieval on clip-level embeddings, 2) re-ranking and filtering of candidates for alignability using DRAQ on contextualized per-frame features, and 3) warping of the best match using DTW.

\section{Evaluating Aligned Video Retrieval}\label{sec:eval}

We propose a protocol to evaluate aligned video retrieval methods. 
Existing video alignment benchmarks do not tackle the problem of \emph{identifying} alignable video pairs and instead build on existing video datasets with well-behaved, alignable action classes like PennAction, where alignability for videos of the same action is assumed. 
Prior works typically report alignment performance via several proxy tasks involving the phase labels, such as phase classification via learned predictors or frame retrieval.
While we also use PennAction and the associated temporal action-phase labels to evaluate our alignment component, we propose a more direct way to measure alignment quality via phase agreement of aligned videos. 
Furthermore, we introduce additional benchmarks for candidate retrieval reranking methods like our proposed alignability indicator, DRAQ.

\noindent \textbf{Problems in Existing Alignment Benchmarks.}
Established video alignment protocols on PennAction primarily consider proxy metrics, such as action phase classification or Kendell Tau, for measuring alignment performance. 
We observe in our experiments that such metrics can to a large extent be gamed, provided sufficient knowledge of each frame's position in the video is encoded in the frame features. 
For example, we find that a BYOL \cite{grill2020bootstrap} frame encoder combined with a temporal Transformer processing the frame embeddings (an architecture analogous to SotA methods like CARL~\cite{chen2022frame}) achieves very high performance \emph{even with a random initialization of the Transformer}.  
We suspect the position encoding's influence on the frame embeddings is the reason for this phenomenon. 
As a result, such proxy tasks might not provide a good indication of real-world alignment performance.

Instead, for datasets like PennAction with dense phase labels, we propose to evaluate the alignment directly by computing phase label agreement after alignment. 
Concretely, we take pairs of videos, compute their temporal alignment according to DTW on the extracted frame features, and report the average agreement of the phase labels after alignment. 
We term this metric Aligned Phase Agreement (APA).

\begin{SCfigure}[50][t]
    \centering
    \sidecaptionvpos{figure}{c} %
    \includegraphics[width=0.47\linewidth, trim=2mm 0mm 3mm 3mm, clip]{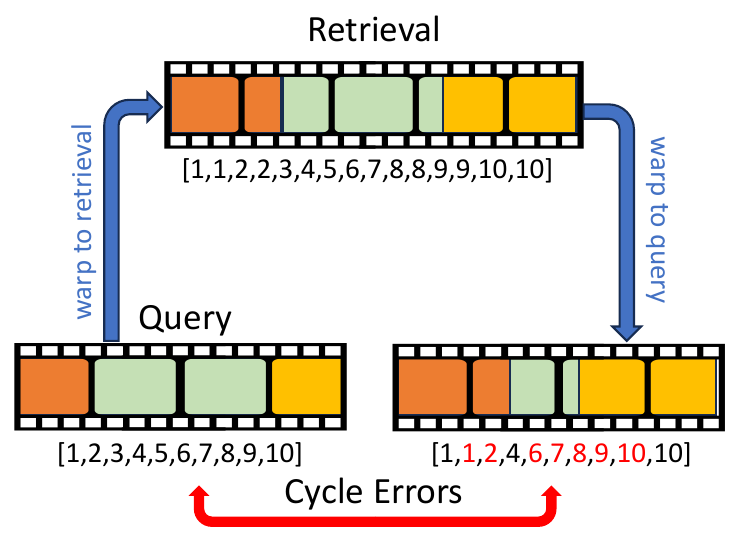}
    \caption{
    \textbf{AVR evaluation via Cycle-Consistency.} 
    We illustrate the use of consistency errors to measure aligned video retrieval performance. A query video (bottom left), along with phase labels (colored regions) and frame indices (below the video), is warped to the top retrieval video (top). 
    The aligned labels and frame indices are then warped back to the query again to complete the cycle. We then report the Frame Position Error (FPE) and the Cycle Phase Error (CPE) when the query contains phase information. 
    }
    \label{fig:cycle}
\end{SCfigure}

\noindent \textbf{Cycle Consistency for AVR Evaluation.}
Ideally, an AVR benchmark would also consist of videos with dense action phase annotations. 
However, obtaining high-quality phase annotations for large-scale and diverse videos is very costly.
Furthermore, it is difficult to consistently define action phases or key events for general videos, as is required for such an approach.
Instead, we propose to leverage cycle consistency as a proxy for AVR performance. 
Concretely, given a query video, we 1) obtain the top candidate using the AVR model, 2) align the query to the top match and propagate per-frame labels (\eg, position or phase labels) to the match, and 3) cycle back to the query with another alignment and label propagation. 
Note that we perform both warps in this cycle so that the second video is kept unwarped, thus ensuring that the cycle-warped video has the same length as the query. 
This is illustrated in Figure~\ref{fig:cycle}.

We propose to use this cycle consistency in two settings: 1) label propagation where the query contains phase annotations but the retrieved match does not, and 2) the position of each frame in the query is used as a "label" for propagation (no human labeling required).
For scenario 1, we report the Cycle Phase Error (CPE) as the average error in phase labels before and after the cycle, and for scenario 2, we report Frame Position Error (FPE) as the MSE between the original and cycle-warped frame position vector. 
Since scenario 1 with CPE only requires the query to contain phase annotation, the approach easily scales to large retrieval datasets and avoids the need for consistent phase annotation between query and retrieval. 
We leverage these benefits to quantify the performance on more general natural videos by labeling a set of Kinetics validation videos with intuitive phase labels (\ie, choosing characteristic key moments in the videos).

\section{Experiments}\label{sec:exp}

We performed cycle-consistency experiments to demonstrate the effectiveness of our solution for Aligned Video Retrieval (AVR) in Section~\ref{sec:exp_avr}.
Additionally, we report experiments verifying the two key components of our method: 1) frame-level contextualized video feature design (Eq.~\ref{eq:frame_feats}) for video alignment in Section~\ref{sec:exp_cont}, and 2) DRAQ for identifying video pairs with the highest alignment quality in Section~\ref{sec:exp_draq}. 

\subsection{Datasets}
We consider three well-known video datasets in our experiments:

\noindent \textbf{UCF101}~\cite{ucf101} is an action recognition dataset containing 13,320 videos of 101 human actions, which are collected from internet videos. We use split-1 for our experiments, where there are 9,537 training videos and 3,783 test videos. 

\noindent \textbf{PennAction}~\cite{pennaction} containing videos of various sports.
We use the same split as prior video alignment work\cite{dwibedi2019temporal, chen2022frame}, covering 2,106 videos with 13 action classes.
Each video has associated video-level action and frame-level action-phase labels.

\noindent \textbf{Kinetics700}~\cite{kay2017kinetics}, a large-scale dataset with about 650,000 natural videos of 700 diverse action classes from the internet. 
We annotate 91 validation videos with intuitive key-frames to define phase labels for cycle-consistent AVR evaluation.

\begin{table}[t]
\renewcommand{\arraystretch}{1.05}
\caption{\textbf{Cycle Consistency for AVR Evaluation.}
We report cycle consistency errors for cycle-warped phase labels (CPE) and frame positions (FPE) on PennAction,  Kinetics, and between PennAction and UCF101. 
The symbol $\circlearrowleft$ means that the query video and retrieval set are from the same dataset, while $\rightleftarrows$ shows that the query video and retrieval set are from different data sources. 
We show results for AVR candidates obtained with retrieval using clip-level \modelName features and oracle retrieval, which randomly chooses candidates that show the same action as the query. 
The performance of BYOL, CARL, and NMS features, all using our feature contextualization, is reported.
For each case, we show the effectiveness of DRAQ reranking, which is applied to the $k=10$ candidates to choose the top match. }
\centering
\begingroup
\resizebox{\linewidth}{!}{
\setlength{\tabcolsep}{2pt}
\arrayrulecolor{black}
\begin{tabular}{ccc!{\color[rgb]{0.853,0.853,0.853}\vrule}d{2.1}d{1.2}!{\color[rgb]{0.853,0.853,0.853}\vrule}d{3.1}d{3.1}!{\color[rgb]{0.853,0.853,0.853}\vrule}d{2.1}d{1.2}} 
\toprule
\multirow{2}{*}{\textbf{Candidates}} & \multirow{2}{*}{\begin{tabular}[c]{@{}c@{}}\textbf{Alignment}\\\textbf{Features}\end{tabular}} & \multirow{2}{*}{\begin{tabular}[c]{@{}c@{}}\textbf{Reranking}\\\textbf{Metric}\end{tabular}} & \multicolumn{2}{c!{\color[rgb]{0.853,0.853,0.853}\vrule}}{\textbf{PennAction}$\circlearrowleft$} & \multicolumn{2}{c!{\color[rgb]{0.853,0.853,0.853}\vrule}}{\textbf{Penn $\rightleftarrows$ UCF}} & \multicolumn{2}{c}{\textbf{Kinetics}$\circlearrowleft$}              \\ 
  &                                                                                              &                                                                                              & \multicolumn{1}{c}{\textbf{FPE} $\downarrow$}   & \multicolumn{1}{c}{\textbf{CPE} $\downarrow$}                                                    & \multicolumn{1}{c}{\textbf{FPE} $\downarrow$}   & \multicolumn{1}{c}{\textbf{CPE} $\downarrow$}                                                     & \multicolumn{1}{c}{\textbf{FPE} $\downarrow$}   & \multicolumn{1}{c}{\textbf{CPE} $\downarrow$ }  \\ 
\arrayrulecolor{black}\hline

\multirow{6}{*}{NMS~\cite{vtn_ssl}} & \multirow{2}{*}{BYOL~\cite{grill2020bootstrap}}                                                                                           & -                                                                                            &       0.5     &       0.40                                                      & 121.1        & 105.0                                                          &     13.0     &  1.03                              \\
 &                                                                                             & DRAQ                                                                                         &      0.2       &            0.13                                                 &  50.6         &  11.0                                                           &     0.3       &  0.09                              \\ \cline{2-9} 
 & \multirow{2}{*}{CARL~\cite{chen2022frame} }                                                                                          & -                                                                                            &     90.3      &       2.38                                                     & 18.7         & 28.5                                                           & 23.5         & 0.45                               \\
 &                                                                                           & DRAQ                                                                                         &    24.3      &       0.74                                                     &  5.2         &  5.9                                                            & 2.3          & 0.08                               \\ 
\arrayrulecolor{black} \cline{2-9}
 &   \multirow{2}{*}{NMS~\cite{vtn_ssl}}                                                                                            & -                                                                                            & 13.4         & 1.32                                                           &  5.5          & 22.2                                                           & 22.7         & 0.86                               \\
 &                                                                                             & DRAQ                                                                                         &  9.5 & 0.20                                                  &  4.8 &  5.9                                                   & 0.5 & 0.0                      \\ \hline
\multirow{6}{*}{Oracle} & \multirow{2}{*}{BYOL~\cite{grill2020bootstrap} }                                                                                          & -                                                                                            & 50.4         & 4.14                                                           & -     & -                                                         & 7.6          & 0.62                               \\
 &                                                                                             & DRAQ                                                                                         &  7.5          & 0.53                                                           &      -    &       -                                                 & 0.3          & 0.05                               \\ 
 \cline{2-9}
 & \multirow{2}{*}{CARL~\cite{chen2022frame}  }                                                                                         & -                                                                                         & 23.4         & 1.34                                                           &  -         &  -                                                          & 36.4         & 1.04                               \\
 &                                                                                           & DRAQ                                                                                         & 11.2         & 0.36                                                           &       -   &  -                                                          & 1.7          & 0.14                               \\ 
\arrayrulecolor{black} \cline{2-9}
 & \multirow{2}{*}{NMS~\cite{vtn_ssl}}                                                                                            & -                                                                                            & 24.7         & 1.70                                                           &      -     &                          -                                 & 35.3         & 1.08                               \\
 &                                                                                         & DRAQ                                                                                         &  7.8 & 0.33                                                  & - & -                                                   & 0.3 & 0.01                      \\ 
\bottomrule
\end{tabular}
}
\endgroup
\label{table:avrretrieval}
\vspace{-2mm}
\end{table}

\subsection{Alignable Video Retrieval Evaluation via Cycle-Consistency}\label{sec:exp_avr}

We evaluate our full AVR pipeline (Section~\ref{sec:method}) in cycle-consistency protocols as described in Section~\ref{sec:eval} on PennAction, UCF101, and Kinetics700. We report results in Table~\ref{table:avrretrieval}.
For candidate proposals, we perform video retrieval (with clip-level features) as described in Section~\ref{sec:method} using the state-of-the-art self-supervised representations of \cite{vtn_ssl} (dubbed NMS).
On PennAction and Kinetics, we also report results with an Oracle proposal generator, choosing candidates from the same class as the query (not reported for Penn$\rightleftarrows$UCF since action classes differ).
In all cases, query videos are taken from the validation set, and retrieval is performed on the training set of the respective datasets.
We compare several off-the-shelf frame-level features \cite{grill2020bootstrap,chen2022frame,vtn_ssl} in combination with our proposed DRAQ re-ranking (applied to the top-10 candidates) regarding temporal alignment performance.
We apply our feature contextualization to all the features since we found it beneficial in all cases (see also Section~\ref{sec:exp_cont}). 
We report average Frame Position Error (FPE) and Cycle Phase Error (CPE) using the phase annotation of the query video.

We can observe clear improvements for DRAQ re-ranked candidates in all cases, which suggests that DRAQ successfully identified the videos that are better aligned among the candidate set. 
While we do observe a lot of variability between feature models and datasets, contextualized NMS features with DRAQ re-reranking appear to perform best overall. 
\begin{figure}[H]
    \centering
    \includegraphics[width=\textwidth]{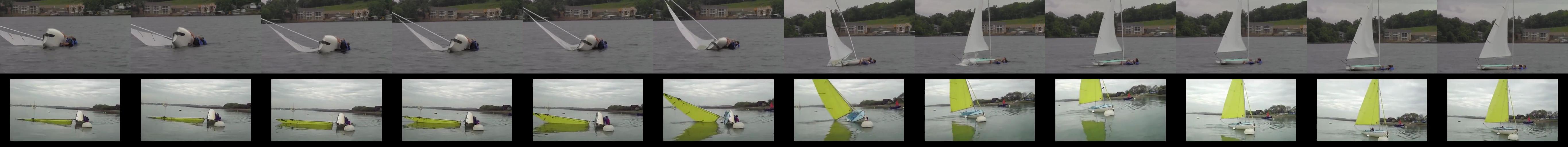}\vspace{2mm}
    \includegraphics[width=\textwidth]{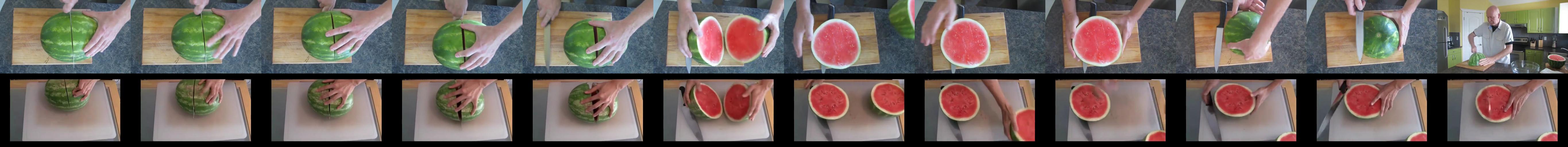}\vspace{2mm}
    \includegraphics[width=\textwidth]{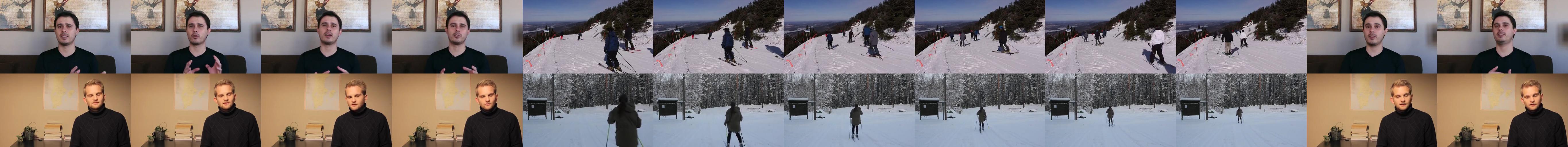}\vspace{2mm}
    \includegraphics[width=\textwidth]{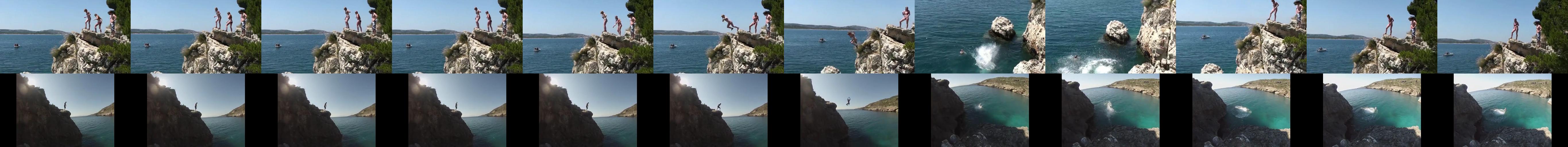}\vspace{2mm}
    \includegraphics[width=\textwidth]{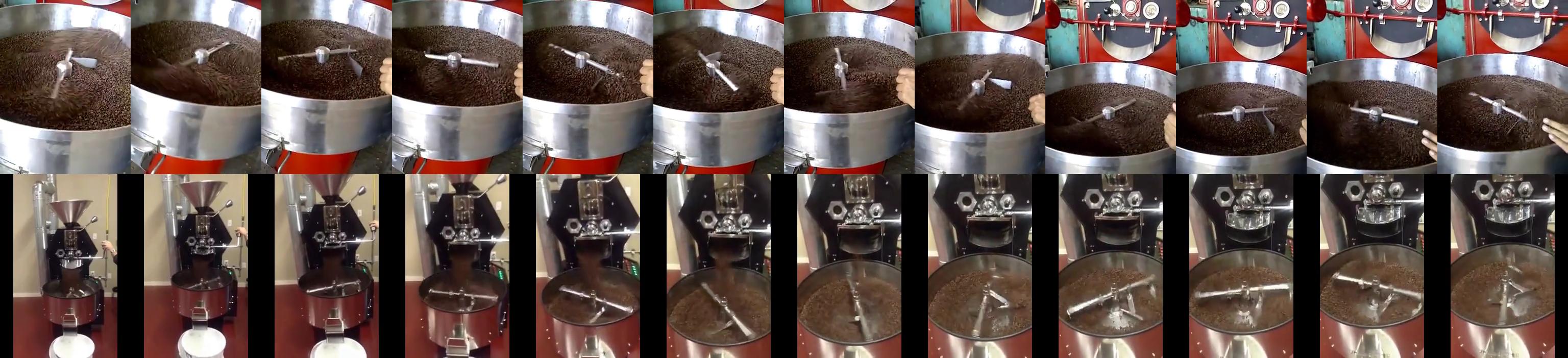}\vspace{2mm}
    \includegraphics[width=\textwidth]{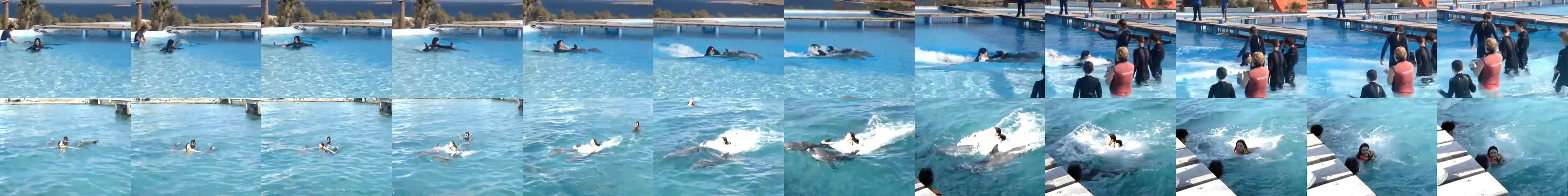}\vspace{2mm}
    \includegraphics[width=\textwidth]{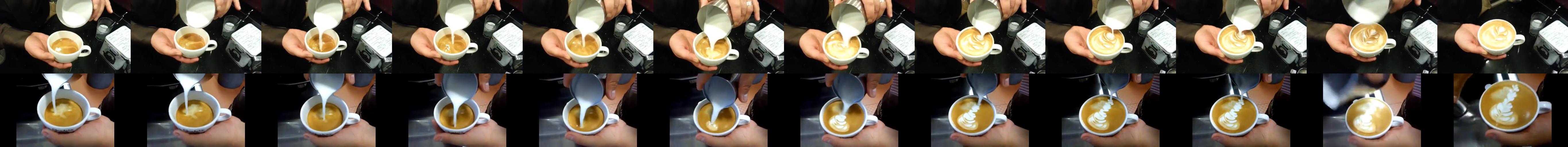}

    \caption{
    \textbf{Qualitative Examples of Aligned Video Retrieval on Kinetics700.}
    The top frame sequence in each row shows the query video (from the validation split), and the bottom sequence shows the aligned retrieval (from the training split) with the lowest DRAQ score among the retrieved candidates. We show results for video pairs with DRAQ$<0.6$, which generally suggests meaningful alignment (zoom in for detail).  
    }
    \label{fig:examples}
\end{figure}
AVR appears to be notably difficult between PennAction and UCF. We hypothesize that this is due to the combination of action class mismatch and the limited size of the retrieval set. 
Qualitative AVR examples on Kinetics are provided in Figure~\ref{fig:examples} and in the Supplementary, where we also provide an expanded table including DTW re-ranking and non-contextualized features.

\subsection{Contextualized Frame-Level Features for Video Alignment}\label{sec:exp_cont}

\begin{table}[t]
\centering
\setlength{\tabcolsep}{4pt}
\begin{tabular}{lcccc} 
\toprule
\textbf{Features}  &  \textbf{+context}  &  \textbf{Avg.}  &   \textbf{Top-DRAQ} \\ 
\midrule
BYOL~\cite{grill2020bootstrap}   &    \xmark     & 0.769     & 0.814                                                                                     \\
CARL~\cite{chen2022frame}   &     \xmark      & 0.826   &     0.856                                                                                 \\
\modelName~\cite{vtn_ssl}  &  \xmark  &  0.832   &  0.868                                                                        \\

\midrule
BYOL~\cite{chen2022frame}  &     \cmark       & 0.801       &  0.821                                                                                \\
CARL~\cite{chen2022frame}  &      \cmark     & 0.827   &   0.854                                                                             \\
\modelName~\cite{vtn_ssl}  &  \cmark     &   0.848   &  0.893                                                                                 \\
\bottomrule
\end{tabular}
\caption{\textbf{Aligned Phase Agreement.} 
We compare frame-level features with and without our contextualization.
We report the Aligned Phase Agreement (APA), \ie, the average agreement of phase labels after warping pairs of videos using DTW. 
Results are computed over the top-10 candidate pairs in our AVR setting (query videos are from PennAction-\texttt{val}, and retrieval is over \texttt{train}).
We report average APA over all candidates and for the top DRAQ re-ranked example per query.}
\label{table:comparison}
\end{table}

To evaluate the improvements due to our proposed contextualized frame-level features, we perform experiments using the action phase annotations on PennAction.
We report average Aligned Phase Agreement accuracy (APA) after DTW alignment for the top-10 retrieval candidate pairs in our AVR setting.  
In Table~\ref{table:comparison}, we compare our contextualized features (+context) to the baseline performance of non-contextualized features from different frame-level feature methods.
We compare features from \modelName~\cite{vtn_ssl} trained through temporal self-supervision against BYOL~\cite{grill2020bootstrap} (a strong self-supervised image representation) and the recent state-of-the-art video alignment method CARL~\cite{chen2022frame}.
We can observe clear improvements with our contextualization (Equation~\ref{eq:frame_feats}) for \modelName and BYOL, which benefit from the added temporal context. 
Since CARL already possesses temporal context due to a longer-range temporal Transformer, we do not observe additional benefits.

\begin{table}[t]
\centering
\begingroup
\resizebox{0.88\linewidth}{!}{
\setlength{\tabcolsep}{4pt}
\begin{tabular}{lcccc} 
\toprule
\textbf{Method} & \textbf{Labels}      & \begin{tabular}[c]{@{}c@{}}\textbf{Phase}\\\textbf{Classification}\\\textbf{(Top-1 Acc.)}\end{tabular} & \begin{tabular}[c]{@{}c@{}}\textbf{Frame}\\\textbf{Retrieval}\\\textbf{Ret@1}\end{tabular} & \begin{tabular}[c]{@{}c@{}}\textbf{Kendall}\\\textbf{Tau}\\$\tau$ \end{tabular}  \\ 
\midrule
TCC~\cite{dwibedi2019temporal}             & Action               & 0.744                                                                                                  & 0.767                                                                                      & 0.641                                                                   \\
GTA~\cite{hadji2021representation}             & Action               & -                                                                                                      & -                                                                                          & 0.748                                                                   \\
LAV~\cite{haresh2021learning}             & Action               & 0.786                                                                                                  & 0.791                                                                                      & 0.684        \\

\hline
TCN~\cite{sermanet2018time}             & None                 & 0.681                                                                                                  & 0.778                                                                                      & 0.542                                                                   \\
SaL~\cite{misra2016shuffle}             & None                 & 0.682                                                                                                  & -                                                                                          & 0.474                                                                   \\
BYOL~\cite{grill2020bootstrap}              & None        &    0.545     &   0.473     &           0.216                                                      \\
CARL~\cite{chen2022frame}             & None                 &     0.931                                                                                         &    0.922                                                                                       & 0.985                                                                   \\

\modelName~\cite{vtn_ssl}      & None                 & 0.799                                                                                        & 0.730                                                                           & 0.397                                                        \\
\hline
BYOL+Transformer             & None                 &      0.863                                                                                            &   0.817                                                                                        & 0.995                                                                   \\

CARL-SW            & None                 &     0.845                                                                                        &     0.830                                                                                       &  0.686                                                                  \\

\midrule
\multicolumn{5}{c}{\textbf{Our Contextualized Frame Features}} \\
\midrule

BYOL + context      & None                 &  0.881\increase{62\%}                                                                                         &  0.782\increase{65\%}                                                                           &  0.776\increase{259\%}                                                          \\
CARL-SW + context        & None                 &     0.889\increase{5\%}                                                                                        &     0.845\increase{2\%}                                                                                       &  0.648\decrease{5\%}                                                                  \\

\modelName + context      & None                 & {0.918}\increase{15\%}                                                                                         & {0.882}\increase{21\%}                                                                             & {0.825}\increase{102\%}                                                          \\
\bottomrule
\end{tabular}
}
\endgroup
\caption{
\textbf{PennAction Benchmarks.}
We demonstrate the effect of our feature contextualization on various frame representations in proxy tasks on PennAction, comparing contextualized features with prior temporal alignment works. These tasks assess the ability to decode temporal information (e.g., action phase) from frame features.
}
\label{table:alignbench}
\end{table}

We also compare contextualized features with prior video alignment methods on various established proxy tasks in Table~\ref{table:alignbench}.
As pointed out in Section~\ref{sec:eval}, we observe shortcuts based on frame position information for models leveraging temporal Transformers (\eg, CARL). 
This is exemplified by the result for BYOL+Transformer, which combines BYOL with a \emph{randomly initialized} untrained temporal Transformer.
This variant represents the initialization of CARL and already outperforms all other prior works (notice the saturated $\tau$ values indicating the shortcut).
We also report results with CARL when applied in a "sliding window" fashion (CARL-SW), similar to how other methods process the videos.
For methods with a limited temporal context (and thus not affected by the position shortcut), we can observe benefits from our feature contextualization. 
Given the observed shortcuts in the proxy tasks of Table~\ref{table:alignbench}, we argue for directly evaluating alignment performance as in Table~\ref{table:comparison} instead.

\begin{SCfigure}[50][t]
    \centering
        \includegraphics[width=0.5\linewidth, valign=t, trim=6mm 2mm 15mm 8mm, clip]{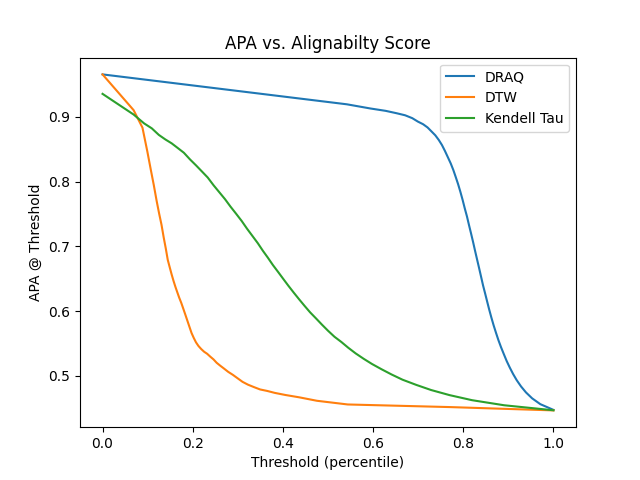}
    \vspace{-8mm}
    \caption{\textbf{DRAQ for Identifying Alignable Videos.}
    We show a plot of the Aligned Phase Agreement (APA) for video pairs with alignment indicators below a given threshold on PennAction. 
    The x-axis corresponds to the percentiles of the respective indicator.
    We compare DRAQ to the optimal DTW cost and Kendell Tau. 
    }
    \label{fig:pa_vs_draq}
    \vspace{7mm}
\end{SCfigure}

\subsection{DRAQ for Measuring Video Alignment Quality}\label{sec:exp_draq}
To verify the ability of DRAQ to identify well-alignable videos among a set of candidates (\eg, obtained through retrieval), we compute the Aligned Phase Agreement (APA) after alignment obtained at different thresholds for DRAQ and alternative alignment quality indicators. 
Videos of the same action class have the same phase labels (assigned to each frame); thus, a high APA indicates that the aligned videos mostly show the same action phase (APA=1 means perfect phase alignment). 
Videos of different actions exhibit an APA of zero. 

We plot the average APA for video pairs that fall below a given threshold for DRAQ and other alignment indicators in Figure~\ref{fig:pa_vs_draq}.
We compare DRAQ against the optimal DTW cost and Kendall $\tau$ (we use $-\tau$ to be consistent with DRAQ and DTW where lower values are better). 
For a fair comparison, we plot the thresholds as percentiles of the respective indicator values. 
The set of video pairs is taken from PennAction and is balanced, \ie, we use the same number of pairs with matching and non-matching action classes.
As the figure shows, pairs with low DRAQ values more consistently achieve high APA. 
This indicates that DRAQ is clearly superior in identifying alignable videos than the alternatives.

\noindent \textbf{Effect of DRAQ on Action Retrieval.}
One of the applications of DRAQ, which is amenable to quantitative analysis, is re-ranking video search results. 
We show results on reranking the top-25 retrievals for PennAction and UCF101 in Table~\ref{table:reranking}.
We measure recall@$k$, where a retrieval is considered correct if the query and retrieval video shows the same action category.  
We observe improved recall in the top retrievals for both datasets with DRAQ re-ranking. 

\begin{table}[t]
\centering
\begingroup
\setlength{\tabcolsep}{4pt}
\begin{tabular}{lcccc} 
\toprule
\multirow{2}{*}{\textbf{Method}} & \multicolumn{2}{c}{\textbf{PennAction}} & \multicolumn{2}{c}{\textbf{UCF101}}  \\
                                 & R@1            & R@10                   & R@1            & R@10                \\ 
\midrule
w/o Reranking                    & 82.09          & 98.76                  & 81.60          & 93.52               \\
DRAQ Reranking                   & {82.40} & {99.17}         & {81.81} & {93.92}      \\
\bottomrule
\end{tabular}
\endgroup
\caption{
\textbf{DRAQ for Action Retrieval Re-Ranking.}
We report recall@$k$ with and without DRAQ re-ranking on PennAction and UCF101. 
DRAQ re-ranking is applied to the top 25 retrievals. }
\label{table:reranking}
\end{table}

\section{Conclusion}
This paper explored the novel task of Aligned Video Retrieval (AVR) to tackle the problem of identifying temporally alignable videos from large collections. 
As a first step towards a solution, we introduced a video alignment score DRAQ, which, given a query video, can help us identify the most alignable videos among a set of candidates. 
In new cycle-consistency benchmarks to measure the performance of AVR, we show that DRAQ, together with carefully designed frame-level features, is able to identify alignable video pairs for general videos with diverse actions. 
With future work, we aim to advance AVR via improved candidate proposals from more sophisticated retrieval. Our work also holds particular interest in retrieval-augmented generation within diffusion models, where from an aligned retrieval video, one can effectively generate corresponding modalities, like audio, for the query video.
We believe that further progress on this task will open up many novel applications for video alignment methods in video editing, discovery, and understanding.

\clearpage  %

\bibliographystyle{splncs04}
\bibliography{main}
\clearpage
\appendix 
\section*{Supplementary Material Overview}
\begin{itemize}
    \item Section~\ref{suppsec:detailedeval}: Detailed Evaluation for Alignable Video Retrieval
    \item Section~\ref{suppsec:qual}: Qualitative Results
    \item Section~\ref{suppsec:discussion}: Discussion on computation and limitations
\end{itemize}

\section{Detailed Evaluation for Alignable Video Retrieval}
\label{suppsec:detailedeval}
We provide an expanded table of ~\supp{Table-1 of our main paper} in Table~\ref{table:full_avr_supp} here.
For all the combinations of alignment features and candidate retrievals, we additionally include the cases of using DTW re-ranking and non-contextualized features. 
We make the following observations:
\begin{itemize}
    \item Within comparable `Alignment Features' and `Reranking Metrics', we observe significant improvements when employing our novel feature contextualization approach. For example, compare \texttt{Row (a1)} with \texttt{Row (d1)}, \texttt{Row (b1)} with \texttt{Row (e1)}, and \texttt{Row (c1)} with \texttt{Row (f1)}, etc.
    \item Within comparable `Alignment Features' and `Context', our proposed DRAQ-based reranking scheme outperforms both the DTW-based reranking and the absence of reranking. For instance, \texttt{Row (c1)} with DRAQ surpasses \texttt{Row (a1)} with no reranking, and \texttt{Row (b1)} with DTW reranking.
\end{itemize}

\section{Qualitative Results}
\label{suppsec:qual}
Qualitative AVR examples of retrieved and aligned video pairs on Kinetics700 are provided in the Supplementary videos. Our qualitative results underscore that our method can effectively achieve alignment within a large-scale dataset, where the top video displays the query and the bottom video showcases the best alignable match retrieved, where both videos are warped with the optimal alignment path $P_{\text{DTW}}$.

For instance, our method precisely matches action phases across various scenarios. This includes watermelon cutting techniques in \texttt{2.mp4}, catch-throw-stance sequences in \texttt{9.mp4}, and the procedural steps within a manufacturing process video in \texttt{16.mp4}. Noteworthy is our method's applicability to a range of general action videos, such as interactions with dolphins in \texttt{6.mp4}, the rotating blade of a coffee machine in \texttt{5.mp4}, and fishing activities in \texttt{15.mp4}.
\begin{table}[H]

\centering
\arrayrulecolor[rgb]{0.753,0.753,0.753}
\begingroup
\resizebox{\linewidth}{!}{
\setlength{\tabcolsep}{1.5pt}
\begin{tabular}{cc|c|c|cc|cc|cc} 
\arrayrulecolor{black}\hline
\multicolumn{1}{l}{}                           & \multirow{2}{*}{\begin{tabular}[c]{@{}c@{}}\textbf{Alignment}\\\textbf{Features}\end{tabular}} & \multirow{2}{*}{\textbf{Context}} & \multirow{2}{*}{\begin{tabular}[c]{@{}c@{}}\textbf{Reranking}\\\textbf{Metric}\end{tabular}} & \multicolumn{2}{c|}{\textbf{PennAction $\circlearrowleft$}} & \multicolumn{2}{c|}{\textbf{Penn $\rightleftarrows$ UCF}} & \multicolumn{2}{c}{\textbf{Kinetics $\circlearrowleft$}}  \\
\multicolumn{1}{l}{}                           &                                                                                                &                                   &                                                                                              & \textbf{FPE} & \textbf{CPE}              & \textbf{FPE} & \textbf{CPE}               & \textbf{FPE} & \textbf{CPE}            \\ 
\hline
\rowcolor[rgb]{0.655,1,1} \multicolumn{1}{l}{} & \multicolumn{9}{c}{\textbf{Video Candidates obtained through NMS~\cite{vtn_ssl} Retrieval}}                                                                                                                                                                                                                                                                                   \\ 
\hline
\texttt{(a1)}                                           & \multirow{6}{*}{BYOL~\cite{grill2020bootstrap}}                                                                          & \multirow{3}{*}{\textcolor[rgb]{0.7,0.7,0.7}{\Large\xmark}}                & -                                                                                            & 125.4        & 3.98                      & 124.5        & 519.87                     & 564.5        & 5.06                    \\
\texttt{(b1)}                                           &                                                                                                &                                   & DTW                                                                                          & 333.7        & 7.95                      & 259.8        & 978.77                     & 1198.4       & 19.2                    \\
\texttt{(c1)}                                           &                                                                                                &                                   & DRAQ                                                                                         & 19.3         & 0.75                      & 72.1         & 283.51                     & 2.3          & 0.24                    \\ 
\arrayrulecolor[rgb]{0.753,0.753,0.753}\cline{3-10}
\texttt{(d1)}                                           &                                                                                                & \multirow{3}{*}{{\Large\cmark}}               & -                                                                                            & 0.54         & 0.4                       & 121.1        & 105.01                     & 13           & 1.03                    \\
\texttt{(e1)}                                           &                                                                                                &                                   & DTW                                                                                          & 0.78         & 0.78                      & 206.6        & 407.08                     & 16.8         & 2.11                    \\
\texttt{(f1)}                                           &                                                                                                &                                   & DRAQ                                                                                         & 0.24         & 0.13                      & 50.6         & 11.03                      & 0.32         & 0.09                    \\ 
\arrayrulecolor{black}\hline
\texttt{(g1)}                                           & \multirow{6}{*}{CARL~\cite{chen2022frame}}                                                                          & \multirow{3}{*}{\textcolor[rgb]{0.7,0.7,0.7}{\Large\xmark}}                & -                                                                                            & 99.8         & 2.34                      & 19.2         & 24.35                      & 40.3         & 0.68                    \\
\texttt{(h1)}                                           &                                                                                                &                                   & DTW                                                                                          & 98.6         & 1.91                      & 11.3         & 25.09                      & 12.2         & 0.34                    \\
\texttt{(i1)}                                           &                                                                                                &                                   & DRAQ                                                                                         & 21.1         & 0.71                      & 21.7         & 10.62                      & 16           & 1.12                    \\
\arrayrulecolor[rgb]{0.753,0.753,0.753}\cline{3-10}
\texttt{(j1)}                                           &                                                                                                & \multirow{3}{*}{{\Large\cmark}}               & -                                                                                            & 90.3         & 2.38                      & 18.7         & 28.49                      & 23.5         & 0.45                    \\
\texttt{(k1)}                                           &                                                                                                &                                   & DTW                                                                                          & 78.3         & 1.68                      & 9.5          & 15.27                      & 9.6          & 0.46                    \\
\texttt{(l1)}                                           &                                                                                                &                                   & DRAQ                                                                                         & 24.3         & 0.74                      & 5.2          & 5.87                       & 2.3          & 0.08                    \\ 
\arrayrulecolor{black}\hline
\texttt{(m1)}                                           & \multirow{6}{*}{NMS~\cite{vtn_ssl}}                                                                           & \multirow{3}{*}{\textcolor[rgb]{0.7,0.7,0.7}{\Large\xmark}}                & -                                                                                            & 72.9         & 2.32                      & 37.2         & 31.9                       & 40           & 1.06                    \\
\texttt{(n1)}                                           &                                                                                                &                                   & DTW                                                                                          & 160.9        & 3.72                      & 51.8         & 26                         & 331.1        & 6.22                    \\
\texttt{(o1)}                                           &                                                                                                &                                   & DRAQ                                                                                         & 83.37        & 0.79                      & 23.5         & 7.82                       & 0.64         & 0.21                    \\ 
\arrayrulecolor[rgb]{0.753,0.753,0.753}\cline{3-10}
\texttt{(p1)}                                           &                                                                                                & \multirow{3}{*}{{\Large\cmark}}               & -                                                                                            & 13.4         & 1.32                      & 5.5          & 22.22                      & 22.7         & 0.86                    \\
\texttt{(q1)}                                           &                                                                                                &                                   & DTW                                                                                          & 14.5         & 1.86                      & 6.7          & 29.77                      & 65           & 1.95                    \\
\texttt{(r1)}                                           &                                                                                                &                                   & DRAQ                                                                                         & 9.5          & 0.2                       & 4.8          & 5.89                       & 0.46         & 0                       \\ 
\arrayrulecolor{black}\hline
\rowcolor[rgb]{0.655,1,1} \multicolumn{1}{l}{} & \multicolumn{9}{c}{\textbf{Video Candidates obtained through Oracle Retrieval}}                                                                                                                                                                                                                                                                                   \\ 
\hline
\texttt{(a2)}                                           & \multirow{6}{*}{BYOL~\cite{grill2020bootstrap}}                                                                          & \multirow{3}{*}{\textcolor[rgb]{0.7,0.7,0.7}{\Large\xmark}}                & -                                                                                            & 197.5        & 5.2                       & -            & -                          & 909.2        & 8.51                    \\
\texttt{(b2)}                                           &                                                                                                &                                   & DTW                                                                                          & 79.5         & 1.19                      & -            & -                          & 980.8        & 15.14                   \\
\texttt{(c2)}                                           &                                                                                                &                                   & DRAQ                                                                                         & 74.5         & 0.85                      & -            & -                          & 6            & 0.04                    \\ 
\arrayrulecolor[rgb]{0.753,0.753,0.753}\cline{3-10}
\texttt{(d2)}                                           &                                                                                                & \multirow{3}{*}{{\Large\cmark}}               & -                                                                                            & 50.4         & 4.14                      & -            & -                          & 7.6          & 0.62                    \\
\texttt{(e2)}                                           &                                                                                                &                                   & DTW                                                                                          & 12.7         & 0.84                      & -            & -                          & 13.4         & 2.1                     \\
\texttt{(f2)}                                           &                                                                                                &                                   & DRAQ                                                                                         & 7.5          & 0.53                      & -            & -                          & 0.3          & 0.05                    \\ 
\arrayrulecolor{black}\hline
\texttt{(g2)}                                           & \multirow{6}{*}{CARL~\cite{chen2022frame}}                                                                         & \multirow{3}{*}{\textcolor[rgb]{0.7,0.7,0.7}{\Large\xmark}}                & -                                                                                            & 41.3         & 1.46                      & -            & -                          & 47.4         & 0.41                    \\
\texttt{(h2)}                                           &                                                                                                &                                   & DTW                                                                                          & 26.6         & 1.11                      & -            & -                          & 4.3          & 0.16                    \\
\texttt{(i2)}                                           &                                                                                                &                                   & DRAQ                                                                                         & 31.6         & 0.41                      & -            & -                          & 2.1          & 0.19                    \\ 
\arrayrulecolor[rgb]{0.753,0.753,0.753}\cline{3-10}
\texttt{(j2)}                                           &                                                                                                & \multirow{3}{*}{{\Large\cmark}}               & -                                                                                            & 23.4         & 1.34                      & -            & -                          & 36.4         & 1.04                    \\
\texttt{(k2)}                                           &                                                                                                &                                   & DTW                                                                                          & 12.5         & 0.93                      & -            & -                          & 4.4          & 0.12                    \\
\texttt{(l2)}                                           &                                                                                                &                                   & DRAQ                                                                                         & 11.2         & 0.36                      & -            & -                          & 1.7          & 0.14                    \\ 
\arrayrulecolor{black}\hline
\texttt{(m2)}                                           & \multirow{6}{*}{NMS~\cite{vtn_ssl}}                                                                           & \multirow{3}{*}{\textcolor[rgb]{0.7,0.7,0.7}{\Large\xmark}}                & -                                                                                            & 88.1         & 3.58                      & -            & -                          & 134.6        & 5.16                    \\
\texttt{(n2)}                                           &                                                                                                &                                   & DTW                                                                                          & 64.2         & 2.97                      & -            & -                          & 157.2        & 3.18                    \\
\texttt{(o2)}                                           &                                                                                                &                                   & DRAQ                                                                                         & 29.5         & 0.87                      & -            & -                          & 0.2          & 0                       \\ 
\arrayrulecolor[rgb]{0.753,0.753,0.753}\cline{3-10}
\texttt{(p2)}                                           &                                                                                                & \multirow{3}{*}{{\Large\cmark}}               & -                                                                                            & 24.7         & 1.7                       & -            & -                          & 35.3         & 1.08                    \\
\texttt{(q2)}                                           &                                                                                                &                                   & DTW                                                                                          & 13.3         & 1.76                      & -            & -                          & 23           & 0.18                    \\
\texttt{(r2)}                                           &                                                                                                &                                   & \multicolumn{1}{c}{DRAQ}                                                                     & 7.8          & 0.33                      & -            & -                          & 0.3          & 0.01                    \\
\arrayrulecolor{black}\hline
\end{tabular}
}
\endgroup
\caption{\textbf{AVR Evaluation.} We report additional results without the proposed contextualized features and with different re-ranking schemes.}
\label{table:full_avr_supp}
\end{table}

\section{Discussion}
\label{suppsec:discussion}
\subsection{Computation Cost, Scalability and Storage}
The first stage of our approach involves retrieving candidate videos through an efficient video retrieval method. We employ established techniques for large-scale search, such as IVFPQ index structures, at this stage of our method. Following this, our approach exhibits constant time $O(1)$ complexity relative to the number of videos in the collection. This is because our DRAQ-based re-ranking computation only applies to a fixed number of top-k candidate videos, thus not impacting scalability. Furthermore, the computational cost of DRAQ (and DTW) is negligible when compared to the computation required for frame-level feature extraction~\cite{vtn_ssl}. The computation time required for DRAQ is just 0.0546\% of the feature extraction time. Finally, the overhead of DRAQ compared to DTW is negligible since it just requires the sampling of random paths through the already computed cost matrix $C$.

Regarding the storage impact of our method, we observe that only the aggregate features must be stored for candidate retrieval (similar to other video retrieval methods).  
The frame-level features of the top-$k$ candidates (for DRAQ) can optionally be computed on the fly, trading off compute for storage cost.

\subsection{Details of NMS~\cite{vtn_ssl}}
NMS enhances a single-frame model by incorporating a temporal Transformer, trained through framewise temporal self-supervision. The primary reason for utilizing~\cite{vtn_ssl} in our method is its ability to deliver state-of-the-art video retrieval results and provide robust frame-wise features. This effectiveness largely stems from its capability to disrupt shortcuts in temporal pretext tasks, thereby promoting accurate frame-level temporal correspondence. For our task, we employ~\cite{vtn_ssl} equipped with a ViT-L backbone, pretrained on Kinetics400.

\section{Limitations}
Our framework assumes that the top matches for alignable videos can be synchronized by identifying clips that are temporally alignable with the query. We presume that within these top matches, all frames are capable of aligning to the query and exhibit temporal monotonicity. However, challenges arise in real-world scenarios where videos may differ in the execution order of action classes. Additionally, some videos may contain extraneous segments that do not match the query (e.g., a video of ``cutting pineapple" might include collecting the fruit from the fridge, which the query does not feature). Despite these complexities, we find that DRAQ is an effective tool for identifying alignable videos in simpler scenarios of non-partial and monotonic alignments. Leveraging a sufficiently expansive search space, this methodology has the potential to enhance the general applicability of video alignment techniques. We identify the resolution of partial matching and non-monotonic alignment as important directions for future extensions.


\end{document}